\documentclass[letterpaper,10pt,conference]{ieeeconf}
\IEEEoverridecommandlockouts
\usepackage[hyphens]{url}
\usepackage{graphicx}
\usepackage{times}
\usepackage[protrusion=false]{microtype}
\usepackage{flushend}
\usepackage{amsmath,amssymb,booktabs,multirow}
\title{\LARGE\bf XS-VLA: Teaching Tiny Vision-Language-Action Models\\with Spatial Supervision and Demonstration Conditioning}
\author{
Iok Tong Lei\textsuperscript{1},
Ying Jie Yap\textsuperscript{1},
Wei Huang\textsuperscript{1},
Qingchen Xie\textsuperscript{1},
Qianzhi Li\textsuperscript{2},
Yujie Zhang\textsuperscript{4},
Xiaolong Liu\textsuperscript{3,*},\\
Zhidong Deng\textsuperscript{1,*}
\thanks{
\textsuperscript{1}
Iok Tong Lei, Qingchen Xie, Wei Huang, Ying Jie Yap, and Zhidong Deng are with the Department of Computer Science and Technology, Tsinghua University, Beijing, China.
}
\thanks{
\textsuperscript{2}
Qianzhi Li is with the National College for Excellent Engineers, Beihang University, Beijing, China.
}
\thanks{
\textsuperscript{3}
Xiaolong Liu is with the Wuxi Dexteroushands Robotic Technology Co.
}
\thanks{
\textsuperscript{4}
Yujie Zhang is an independent researcher.
}
\thanks{
\textsuperscript{*}
Corresponding authors: Xiaolong Liu and Zhidong Deng.
Emails: \texttt{xllau@126.com}, \texttt{michael@mail.tsinghua.edu.cn}.
Primary contact: Iok Tong Lei, \texttt{lixt22@mails.tsinghua.edu.cn}.
}
\thanks{
Project page:
https://999tom.github.io/xs-vla/
}
}
\begin{document}
\maketitle
\thispagestyle{empty}
\pagestyle{empty}

\begin{abstract}
How can richer training supervision improve robot control while keeping the deployed policy compact? We present XS-VLA, a staged training framework using teacher-derived spatial labels and demonstration-conditioned action learning. Coarse-Grained Spatial Distillation (CSD) initializes the backbone through an auxiliary region-label task. Latent Flow Matching (LFM) then conditions an action-space velocity field on a demonstration latent, using KL regularization while jointly optimizing the backbone and action modules. The deployed policy contains 243.99M parameters and operates without the teacher or posterior encoder. XS-VLA achieves 90.25\% average LIBERO success in each of two training seeds, compared with 86.00\% for a SmolVLA-0.25B base trained under our settings. Ablations examine both training stages through matched image pretraining and Huber/MSE controls. On three Mobile ALOHA tasks, average strict success increases from 21.7\% to 65.0\%. These results demonstrate the control utility of auxiliary representation initialization and regularized demonstration-conditioned flow learning for compact VLA~policies.
\end{abstract}

\section{Introduction}
Vision-language-action (VLA) models use pretrained visual-language representations for robot control~\cite{brohan2023rt2,kim2024openvla,octo2023}. Compact architectures such as SmolVLA~\cite{shukor2025smolvla} support deployment under resource constraints. We ask how richer training supervision can improve manipulation with approximately 244M deployed parameters.

Training offers two sources of information beyond ordinary policy inputs: teacher-generated spatial labels and future actions in completed demonstrations. Spatial labels provide an auxiliary task for initializing the backbone, while demonstration actions provide context for learning a control policy. XS-VLA uses both sources during training and transfers their influence through the learned policy parameters.

XS-VLA organizes these objectives into two stages. Coarse-Grained Spatial Distillation (CSD) turns teacher-predicted object centers into region labels for backbone pretraining. Latent Flow Matching (LFM) conditions the velocity field on demonstration latents, regularizes their posterior toward a Gaussian prior, and jointly optimizes the backbone and action modules. At deployment, observations and instructions drive action generation with a fixed latent. Neither the teacher nor the posterior encoder is executed, and region predictions are not required.

Spatial representations motivate CSD~\cite{chen2024spatialvlm,qu2025spatialvla,zheng2024tracevla}; ACT and generative action policies motivate LFM~\cite{zhao2023act,chi2023diffusion,black2026pi0}. Building on SmolVLA, XS-VLA brings these training signals into a compact controller. We evaluate their utility through LIBERO configuration comparisons and physical manipulation, and use spatial diagnostics to characterize the CSD~initialization.

Our contributions are:
\begin{enumerate}
\item \textbf{A staged training framework for compact control.} XS-VLA uses coarse spatial supervision for representation initialization and a KL-regularized demonstration latent for flow learning, yielding a 243.99M-parameter deployed policy without its teacher or posterior encoder.
\item \textbf{Control evidence for the training design.} Configuration ablations evaluate the initialization and regularized latent branch, with Huber and MSE controls. The full framework reaches 90.25\% LIBERO success in two training seeds, versus 86.00\% for our retrained base.
\item \textbf{Physical manipulation and deployment measurements.} Average strict success on three Mobile ALOHA tasks rises from 21.7\% to 65.0\%. We report trial counts, uncertainty intervals, GPU memory and latency, and qualitative deployments on OpenARM and~PiPER.
\end{enumerate}

\begin{figure*}[t]
\vspace*{4pt}
\centering
\includegraphics[width=\textwidth]{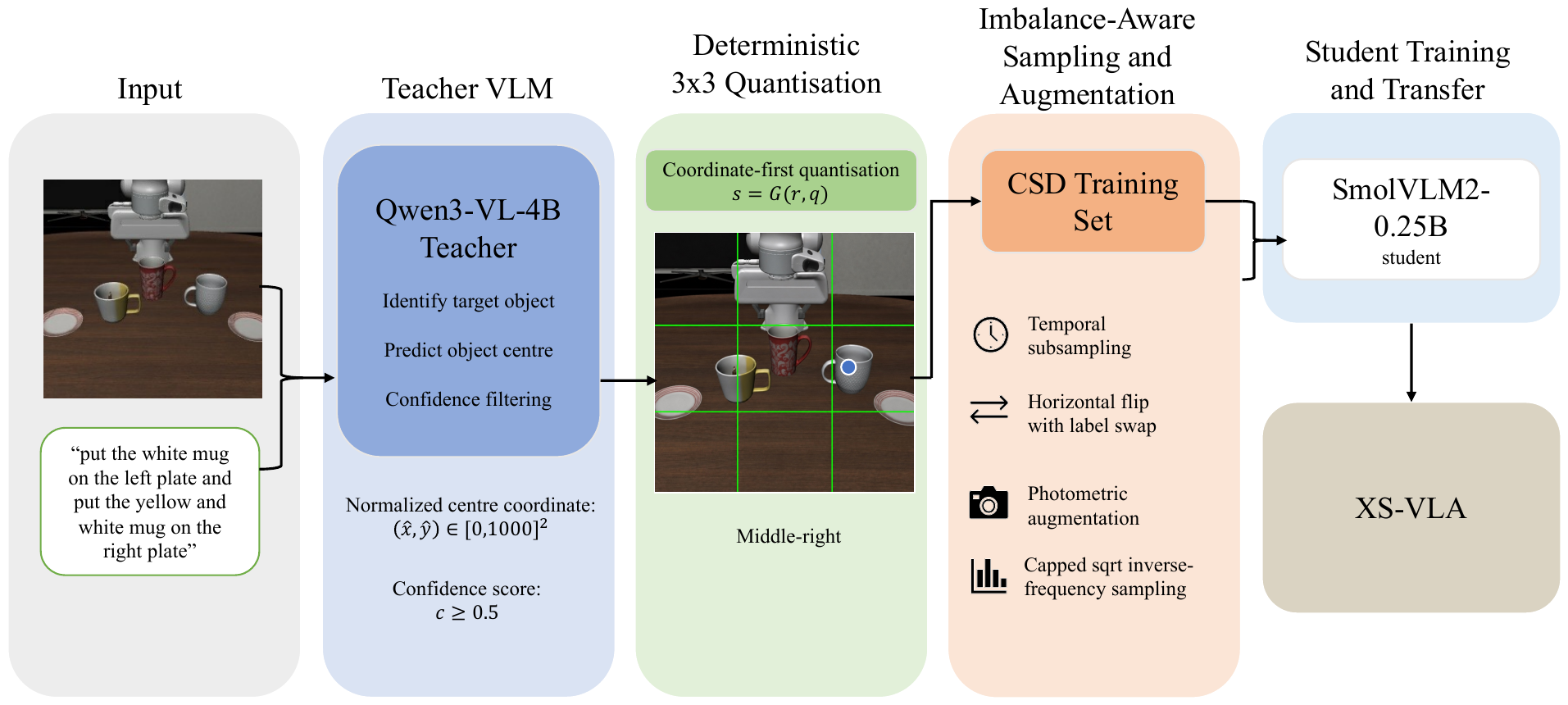}
\caption{Coarse-Grained Spatial Distillation for representation initialization. The teacher predicts a target center and confidence from the original image and instruction. Accepted centers are quantized into nine labels for an auxiliary student task. The overlay illustrates the quantization grid; the teacher processes unmarked images. The learned backbone parameters initialize the policy.}
\label{figure1}
\end{figure*}

\section{Related Work}
\textbf{Compact visual-language control.}
RT-2~\cite{brohan2023rt2}, OpenVLA~\cite{kim2024openvla}, and Octo~\cite{octo2023} demonstrate how pretrained representations support robot policies. SmolVLM2~\cite{padmanabhan2024smolvlm}, MobileVLM~\cite{chu2023mobilevlm_full}, and TinyLLaVA~\cite{wu2024tinyllava_full} target smaller visual-language models. SmolVLA~\cite{shukor2025smolvla} combines a compact visual-language backbone with a flow-based action expert. XS-VLA extends this architecture with auxiliary representation initialization, demonstration-conditioned flow learning, and joint backbone--action training.

\textbf{Spatial supervision and knowledge transfer.}
SpatialVLM~\cite{chen2024spatialvlm} and SpatialVLA~\cite{qu2025spatialvla} study spatial representations; TraceVLA~\cite{zheng2024tracevla} uses visual traces, and ThinkAct~\cite{huang2025thinkact} studies visual latent planning. Task-specific distillation can transfer VLM knowledge to smaller networks~\cite{jang2025vl2lite}. CSD converts teacher coordinates into coarse symbolic targets for auxiliary pretraining. Its $3\times3$ vocabulary expresses spatial cues through the backbone's text-generation interface. The resulting weights initialize the control backbone, and subsequent policy training adapts them to action prediction.

\textbf{Latent-conditioned action generation.}
ACT~\cite{zhao2023act} reports benefits from CVAE training on human demonstrations, motivating latent conditioning for action learning. Diffusion Policy~\cite{chi2023diffusion} and flow matching~\cite{lipman2023flow,black2026pi0} model continuous action distributions. LFM uses a demonstration latent to condition an action-space velocity field during training. Integrating the learned field generates action chunks; deployment uses the prior-mean latent and bypasses the posterior encoder.

\section{Method}
\label{sec:method}
XS-VLA first learns a backbone initialization with an auxiliary region-label objective, updating language and cross-modal parameters with the vision encoder frozen. It then jointly optimizes the backbone, latent encoder, and flow expert on robot demonstrations (Figs.~\ref{figure1}--\ref{figure2}). The resulting policy contains 243.99M deployed parameters, including the action expert (Section~\ref{sec:deployment_profile}).

\subsection{Coarse-Grained Spatial Distillation}
\textbf{Teacher supervision.}
For an RGB observation $I_t$ and instruction $\ell$, Qwen3-VL-4B~\cite{bai2025qwen3vltechnicalreport} predicts the normalized target center $\hat{\mathbf p}_t=(\hat x_t,\hat y_t)\in[0,1000]^2$ and confidence $c_t\in[0,1]$. The teacher receives no markers, bounding boxes, keypoints, or ground-truth coordinates. It is prompted to identify the physical object to manipulate, excluding the robot, gripper, and destination container unless they are themselves the target. For sequential instructions, it selects the object associated with the earliest unfinished operation visible in the image. Teacher decoding uses temperature zero, a fixed seed, and a constrained JSON schema. We retain labels with $c_t\geq0.5$.

We quantize the accepted center into a $3\times3$ grid:
\begin{equation}
r_t=\min\!\left(\left\lfloor\frac{3\hat y_t}{1001}\right\rfloor,2\right),\quad
q_t=\min\!\left(\left\lfloor\frac{3\hat x_t}{1001}\right\rfloor,2\right).
\end{equation}
A deterministic map assigns $s_t=\mathcal G(r_t,q_t)$ to a vocabulary $\mathcal S$ of nine labels. Rows denote top, middle, and bottom; columns denote left, center, and right, with the middle cell labeled \texttt{center}. Coordinates within the same cell receive the same target, reducing sensitivity to small variations in the teacher's predictions within that region.

Region-label prediction supplies spatial supervision before action learning. It uses the backbone's text-generation interface to learn an initialization that is subsequently adapted through robot demonstrations. The auxiliary task thus influences control through the transferred parameters.

\textbf{Dataset and sampling.}
The procedure yields 27,721 accepted frame-level labels from 100 LIBERO demonstration episodes~\cite{libero}. Episodes 0--79 are used for CSD training and episodes 80--99 for held-out teacher-label validation. This episode split prevents adjacent frames from the same trajectory from crossing the CSD split. Retaining every fifth training frame yields 4,510 examples. To reduce class imbalance, we sample with capped square-root inverse-frequency weights
\begin{equation}
w_s=\min\!\left(\sqrt{n_{\max}/n_s},10\right),
\end{equation}
where $n_s$ counts retained examples in a nonempty class and $n_{\max}$ is the largest class count. Horizontal reflection is applied with probability 0.5 and exchanges left/right region labels; center-column labels are unchanged. Photometric augmentation varies brightness, contrast, saturation, hue, and sharpness. Translation and rotation are excluded because coarse labels cannot be updated precisely for these~transformations.

\begin{figure*}[t]
\centering
\includegraphics[width=\textwidth]{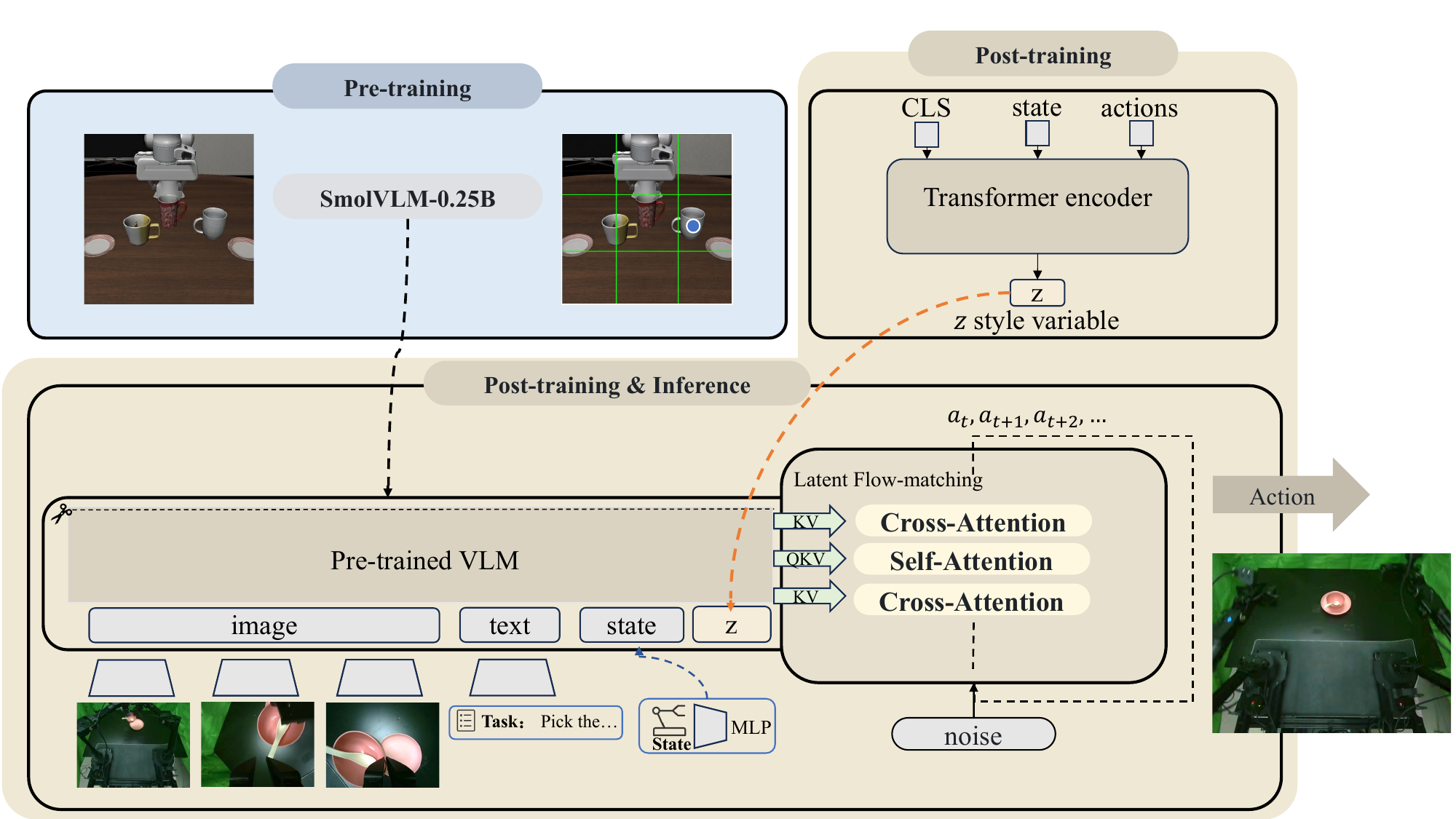}
\caption{XS-VLA architecture. The CSD-initialized backbone supplies observation context to the flow expert. A Transformer encodes state and actions into a demonstration latent, regularized toward a Gaussian prior. Inference bypasses the encoder and uses $\mathbf z=\mathbf0$. The spatial overlay denotes auxiliary pretraining supervision.}
\label{figure2}
\end{figure*}

\textbf{Student training and transfer.}
The student starts from \texttt{SmolVLM2-256M-Video-Instruct}, abbreviated as SmolVLM2-0.25B in Figs.~\ref{figure1}--\ref{figure2}. We refer to our baseline built on this backbone as SmolVLA-0.25B base. These names indicate the pretrained backbone's nominal model scale; XS-VLA's complete deployed policy contains 243.99M parameters, including the action expert. Given $I_t$, $\ell$, and a prompt requesting exactly one allowed region label, it predicts the tokenized target $s_{t,1:L}$ using teacher forcing:
\begin{equation}
\mathcal L_{\mathrm{CSD}}=-\frac1L\sum_{j=1}^L
\log p_\theta(s_{t,j}\mid I_t,\ell,s_{t,<j}).
\end{equation}
CSD trains for 14,000 steps with batch size four using AdamW, a learning rate of $10^{-4}$, weight decay of $10^{-2}$, gradient-norm clipping at 10, and token-level label smoothing of 0.1. The vision encoder is frozen while language and cross-modal parameters are optimized; the action expert is absent. The step-14,000 checkpoint initializes policy training.

Compatible CSD parameters initialize XS-VLA's 16-layer backbone, and the language modeling head is discarded. Policy training then jointly updates the vision encoder, language backbone, action expert, and latent encoder to adapt the initialization to manipulation.

\subsection{Latent Flow Matching}
LFM introduces a latent summary of demonstration actions as context for the flow-matching velocity field. Motivated by ACT's CVAE results~\cite{zhao2023act}, we learn a posterior with KL regularization and use the prior mean during deployment.

\textbf{Inferring the demonstration latent.}
Let $\mathbf o_t$ contain multi-view images, proprioceptive state $\mathbf c_t$, and instruction $\ell$. The policy predicts a chunk $A_t=(\mathbf a_t,\ldots,\mathbf a_{t+K-1})$ of $K=50$ actions. A CVAE-style encoder~\cite{sohn2015learning,zhao2023act} defines a diagonal Gaussian posterior
\begin{equation}
q_\phi(\mathbf z\mid A_t,\mathbf c_t)
=\mathcal N(\boldsymbol\mu,\operatorname{diag}(\boldsymbol\sigma^2)),
\quad p(\mathbf z)=\mathcal N(\mathbf0,\mathbf I),
\end{equation}
where $\mathbf z\in\mathbb R^{32}$. A Transformer encodes the state and action sequence using 256-dimensional token embeddings. Its \texttt{[CLS]} representation predicts $\boldsymbol\mu$ and $\log\boldsymbol\sigma^2$. Reparameterization gives $\mathbf z=\boldsymbol\mu+\boldsymbol\sigma\odot\boldsymbol\epsilon_z$, where $\boldsymbol\epsilon_z\sim\mathcal N(\mathbf0,\mathbf I)$.

\textbf{Conditioning the velocity field.}
The SmolVLA-based expert~\cite{shukor2025smolvla} receives image-language features and projected state as a multimodal prefix. We append the projected latent $\mathbf z$ as an additional conditioning token. Noisy action tokens and flow-time embeddings form the suffix. Cross-attention to the prefix, with periodic joint self-attention, makes the demonstration latent available to velocity prediction throughout flow training.

Here, $\tau=1$ denotes noise and $\tau=0$ denotes demonstrated actions. With $\boldsymbol\epsilon_a\sim\mathcal N(\mathbf0,\mathbf I)$ and $\tau\sim\mathrm{Beta}(1.5,1.0)$, the interpolation and target velocity are
\begin{equation}
A_t^\tau=(1-\tau)A_t+\tau\boldsymbol\epsilon_a,
\qquad \mathbf u=\boldsymbol\epsilon_a-A_t.
\end{equation}
The full model uses Huber regression with threshold $\delta=1.0$, equivalent to Smooth-L1 with $\beta=1.0$:
\begin{equation}
\mathcal L_{\mathrm{FM}}=
\mathbb E\!\left[H_1\!\left(\mathbf v_\theta(A_t^\tau,\mathbf o_t,\mathbf z,\tau)-\mathbf u\right)\right].
\end{equation}
The expectation covers training examples, posterior latents, flow times, and action noise. The complete objective is
\begin{align}
\mathcal L_{\mathrm{KL}}&=D_{\mathrm{KL}}(q_\phi(\mathbf z\mid A_t,\mathbf c_t)\Vert p(\mathbf z)),\\
\mathcal L&=\mathcal L_{\mathrm{FM}}+\lambda_{\mathrm{KL}}(n)\mathcal L_{\mathrm{KL}}.
\end{align}
The flow loss has weight 1.0 with no additional scaling. At policy-training step $n$, $\lambda_{\mathrm{KL}}(n)=10\min(n/10{,}000,1)$, increasing linearly from zero to 10.0 over the first 10,000 steps and remaining at 10.0 thereafter.

\textbf{Roles of conditioning and regularization.}
The posterior latent supplies demonstration-specific context to velocity learning, while KL encourages its distribution to match the Gaussian prior. Huber specifies the penalty on velocity residuals. Training therefore combines a conditioned velocity objective with posterior regularization; deployment uses the prior mean in place of the demonstration-conditioned latent.

\textbf{Deployment.}
Inference starts from standard Gaussian action noise at $\tau=1$ and integrates to $\tau=0$ using 10 explicit Euler steps with $\Delta\tau=-0.1$. Each step updates $A_t^{\tau+\Delta\tau}=A_t^\tau+\Delta\tau\,\mathbf v_\theta(A_t^\tau,\mathbf o_t,\mathbf0,\tau)$. The posterior encoder is bypassed, and $\mathbf z=\mathbf0$ is fixed across queries. Action sampling uses the initial Gaussian noise.

\section{Experiments}
LIBERO ablations examine representation initialization and latent training, while Mobile ALOHA tests physical manipulation. We also profile deployment costs and characterize the CSD checkpoint through spatial diagnostics.

\subsection{LIBERO Setup and Main Results}
We train XS-VLA for 160,000 steps on LeRobot-formatted LIBERO demonstrations using AdamW with global batch size 32, weight decay $10^{-10}$, and $(\beta_1,\beta_2)=(0.9,0.95)$. The learning rate warms up linearly to $10^{-4}$ over the first 1,000 global steps, cosine-decays to $2.5\times10^{-6}$ at global step 30,000, and remains at that value through step 160,000. The cosine schedule uses the absolute global step without resetting after~warmup.

Evaluation uses 10 episodes per task with a fixed seed across Spatial, Object, Goal, and Long. Seven-dimensional LIBERO actions are padded to 32. XS-VLA generates 50-action chunks using 10 Euler steps. The execution horizon is 50 in the training configuration but overridden to 10 for LIBERO evaluation: only the first 10 actions are executed before replanning.

\textbf{Baselines and result sources.}
Table~\ref{tab:main_results} places XS-VLA alongside published policy systems under their original protocols. The three SmolVLA entries, together with Diffusion Policy and Octo scores, are taken directly from Shukor et al.~\cite{shukor2025smolvla}; other external scores follow their respective publications. These results provide benchmark context. Table~\ref{tab:ablation_libero} compares our training configurations using a separate SmolVLA-0.25B base trained under our settings. Unlike the frozen-VLM recipe used for the published SmolVLA results, our 86.00\% LIBERO baseline unfreezes the VLM during policy training, as does XS-VLA.

\textbf{Reporting precision.}
LIBERO success rates in Tables~\ref{tab:main_results} and~\ref{tab:ablation_libero} are rounded to one decimal place for compact presentation. Detailed discussions of our LIBERO runs report overall success rates to two decimal places; for example, 361 successes out of 400 episodes yield 90.25\%, displayed as 90.3\% in the tables. The same one-decimal format is used when discussing the published-result comparison. These representations refer to the same underlying results, and rounding is applied only for display.

\begin{table*}[t]
\centering
\caption{System-level LIBERO success rates (\%). Published results retain their original protocols and parameter-size labels. ``Pretr.'' denotes action-trajectory pretraining. XS-VLA has 243.99M deployed parameters, including the action expert.}
\label{tab:main_results}
\small
\begin{tabular}{lcccccc}
\toprule
Model & Pretr. & Spatial & Object & Goal & Long & Avg.\\
\midrule
\multicolumn{7}{l}{\emph{Published results}}\\
Diffusion Policy~\cite{chi2023diffusion} & No & 78.3 & 92.5 & 68.3 & 50.5 & 72.4\\
TraceVLA (7B)~\cite{zheng2024tracevla} & Yes & 84.6 & 85.2 & 75.1 & 54.1 & 74.8\\
ThinkAct (7B)~\cite{huang2025thinkact} & Yes & 88.3 & 91.4 & 87.1 & 70.9 & 84.4\\
FPC-VLA (7B)~\cite{fpcvla2025} & Yes & 86.2 & 87.0 & 92.0 & 82.2 & 86.9\\
OpenVLA (7B)~\cite{kim2024openvla} & Yes & 84.7 & 88.4 & 79.2 & 53.7 & 76.5\\
SpatialVLA (4B)~\cite{qu2025spatialvla} & Yes & 88.2 & 89.9 & 78.6 & 55.5 & 78.1\\
Octo (0.09B)~\cite{octo2023} & Yes & 78.9 & 85.7 & 84.6 & 51.1 & 75.1\\
SmolVLA (2.25B)~\cite{shukor2025smolvla} & No & 93.0 & 94.0 & 91.0 & 77.0 & 88.8\\
SmolVLA (0.45B)~\cite{shukor2025smolvla} & No & 90.0 & 96.0 & 92.0 & 71.0 & 87.3\\
SmolVLA (0.24B)~\cite{shukor2025smolvla} & No & 87.0 & 93.0 & 88.0 & 63.0 & 82.8\\
\midrule
\multicolumn{7}{l}{\emph{Evaluated in this work}}\\
\textbf{XS-VLA (ours)} & No & 93.0 & 91.0 & 88.0 & 89.0 & \textbf{90.3}\\
\bottomrule
\end{tabular}
\end{table*}

XS-VLA exceeds published SmolVLA-0.24B success by 7.5 points (90.3\% versus 82.8\%), improving Spatial and Long with lower Object and unchanged Goal success.

Tables~\ref{tab:main_results} and~\ref{tab:ablation_libero} report our original runs. The full-model training-seed check below retains the same evaluation seed and protocol.

\subsection{Ablation and Configuration Study}
\begin{table*}[t]
\vspace*{6pt}
\centering
\caption{LIBERO configurations and loss controls (one training run per row). Image pretraining uses generic text or CSD labels. LFM uses posterior conditioning, KL regularization, and Huber regression; the base uses MSE. Overall averages four suites (400 episodes). Avg. steps counts environment steps per episode, including the effects of success and termination.}
\label{tab:ablation_libero}
\small
\setlength{\tabcolsep}{4pt}
\begin{tabular}{lccccrrrrr}
\toprule
\multirow{2}{*}{\textbf{Variant}} & \multirow{2}{*}{\textbf{CSD}}
& \multirow{2}{*}{\shortstack{\textbf{Image}\\\textbf{pretr.}}}
& \multirow{2}{*}{\shortstack{\textbf{Latent}\\\textbf{branch}}}
& \multicolumn{5}{c}{\textbf{Success rate (\%)}}
& \multirow{2}{*}{\shortstack{\textbf{Avg. steps}\\\textbf{/ episode}}}\\
\cmidrule(lr){5-9}
& & & & Spatial & Object & Goal & Long & Overall & \\
\midrule
SmolVLA-0.25B base & No & No & No & 93.0 & 88.0 & 83.0 & 80.0 & 86.0 & 180.65\\
CSD only & Yes & Yes & No & 82.0 & 99.0 & 91.0 & 83.0 & 88.8 & 177.90\\
LFM-only & No & No & Yes & 93.0 & 90.0 & 85.0 & 82.0 & 87.5 & 177.51\\
Image-only + LFM & No & Yes & Yes & 90.0 & 94.0 & 82.0 & 89.0 & 88.8 & 173.85\\
\midrule
CSD + Huber (no latent) & Yes & Yes & No & 90.0 & 89.0 & 82.0 & 85.0 & 86.5 & 177.73\\
XS-VLA (MSE) & Yes & Yes & Yes & 94.0 & 93.0 & 88.0 & 84.0 & 89.8 & 173.82\\
\textbf{CSD + LFM (XS-VLA)} & Yes & Yes & Yes & 93.0 & 91.0 & 88.0 & 89.0 & \textbf{90.3} & 172.54\\
\bottomrule
\end{tabular}
\end{table*}
\textbf{Training configurations.}
Our SmolVLA-0.25B base uses MSE without CSD or latent conditioning and achieves 86.00\% success (344/400). CSD-only reaches 88.75\% (+2.75 points), while LFM-only reaches 87.50\% (+1.50 points) (Table~\ref{tab:ablation_libero}).

\textbf{Combining CSD and LFM.}
The full framework reaches 90.25\%, exceeding the base by 4.25 points, CSD-only by 1.50 points, and LFM-only by 2.75 points. On Long, success rises from the base's 80.0\% to 89.0\%. XS-VLA matches or exceeds LFM-only across suites, although CSD-only performs better on Object and Goal. The combined configuration benefits aggregate control performance, with task-specific tradeoffs.

\textbf{Effect of the regularized latent branch.}
With CSD and Huber, full LFM reaches 90.25\% (361/400), versus 86.50\% (346/400) without the latent branch: a 3.75-point advantage. Here, the added branch comprises both posterior conditioning and KL regularization. The gain supports this combined training mechanism.

\textbf{Huber and MSE with latent training.}
XS-VLA (MSE) replaces only the Huber velocity-regression loss with MSE, retaining CSD, latent conditioning, and the KL warmup schedule. After the 10,000-step linear KL warmup from zero to 10, its objective is $\mathcal L_{\mathrm{total}}=\mathcal L_{\mathrm{FM}}^{\mathrm{MSE}}+10D_{\mathrm{KL}}(q_\phi\Vert p)$.

XS-VLA (MSE) reaches 89.75\% (359/400), 0.50 points (two episodes) below full LFM with Huber. Both losses yield strong control performance. Huber raises Long from 84\% to 89\%, while Spatial and Object decline. Together with the no-latent Huber control, this supports the regularized latent branch's contribution beyond the choice of Huber alone.

\textbf{Spatial labels versus image-only pretraining.}
Image-only pretraining matches CSD's LIBERO image split, sampling, augmentations, initialization, frozen modules, optimizer, batch size, and 14,000-step schedule. Its supervision is the fixed sentence \texttt{A manipulation task.}, without spatial labels. Both variants then receive 160,000 policy-training steps with LFM. Image-only reaches 88.75\% (355/400), versus 90.25\% with CSD: a 1.50-point advantage for CSD initialization beyond matched image exposure and pretraining steps. Both score 89.0\% on Long.

\textbf{Training-seed check.}
Changing only the training seed to 1004, full XS-VLA at 160,000 steps scores 93/93/85/90\% on Spatial/Object/Goal/Long and 90.25\% overall (361/400). The original run scores 93/91/88/89\%, also 90.25\%. Two-seed suite means are 93.0/92.0/86.5/89.5\%. The aggregate score is unchanged across these two training seeds, although suite outcomes vary. Other configurations retain one training run~each.

\subsection{Real-World Evaluation on Mobile ALOHA}
\label{sec:realworld}
\begin{figure*}[t]
\centering
\includegraphics[width=0.92\textwidth]{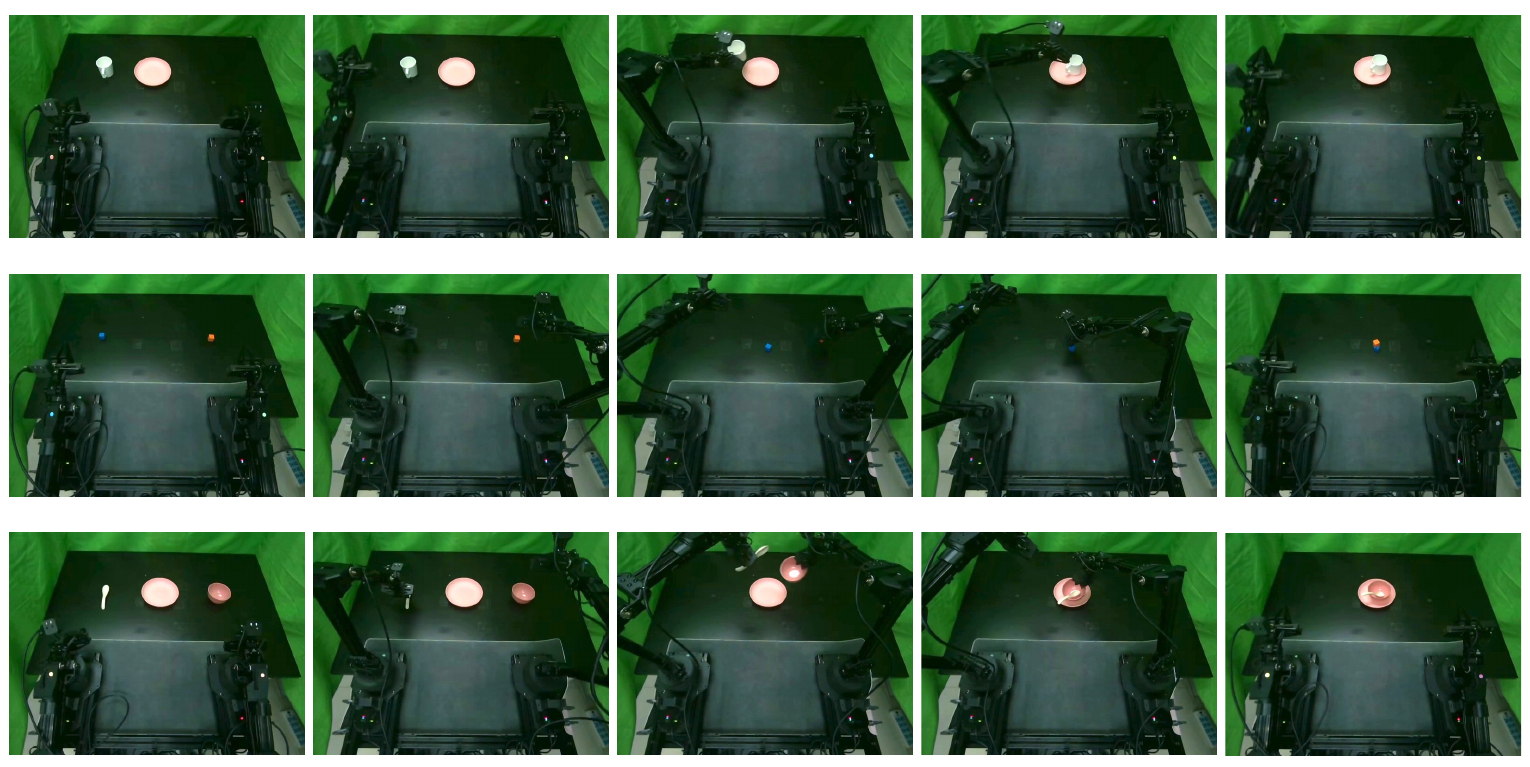}
\caption{Mobile ALOHA demonstrations: cup placing (top), dual-arm block stacking (middle), and sequential spoon--bowl coordination (bottom). Each task uses 100 demonstrations from three operators and 20 evaluation trials per model.}
\label{fig:mobile_aloha_demo}
\end{figure*}

\begin{table*}[t]
\centering
\caption{Mobile ALOHA strict success: percentage (successful trials / 20), followed by a 95\% Wilson interval computed from the counts assuming independent Bernoulli trials. Stacking score is partial credit out of 40. Average uses the three strict success rates.}
\label{tab:real_world_results_enriched}
\small
\begin{tabular}{lccccc}
\toprule
\multirow{2}{*}{Model} & \multirow{2}{*}{Cup placing}
& \multicolumn{2}{c}{Block stacking}
& \multirow{2}{*}{Sequential} & \multirow{2}{*}{Average}\\
\cmidrule(lr){3-4}
& & Strict success & Stack score & & \\
\midrule
SmolVLA-0.25B base & 45.0\% (9/20) & 0.0\% (0/20) & 3/40 & 20.0\% (4/20) & 21.7\%\\
95\% interval & [25.8, 65.8] & [0.0, 16.1] & --- & [8.1, 41.6] & ---\\
\midrule
XS-VLA & \textbf{90.0\% (18/20)} & \textbf{30.0\% (6/20)} & \textbf{20/40} & \textbf{75.0\% (15/20)} & \textbf{65.0\%}\\
95\% interval & [69.9, 97.2] & [14.5, 51.9] & --- & [53.1, 88.8] & ---\\
\bottomrule
\end{tabular}
\end{table*}

\textbf{Tasks and data.}
We evaluate three tasks on Mobile ALOHA~\cite{fu2024mobilealohalearningbimanual} (Fig.~\ref{fig:mobile_aloha_demo}). Cup placing tests single-arm placement on a target plate. Dual-arm block stacking requires a stable stack of small wooden blocks with a $25\,\mathrm{mm}\times25\,\mathrm{mm}$ footprint. Sequential coordination requires the left arm to place a spoon into a bowl held by the right arm, followed by the right arm placing the bowl and spoon on a plate.

For each task, three human operators collect 100 demonstrations with variation in approach, grasp pose, speed, and coordination. XS-VLA initializes its backbone directly from the 14,000-step CSD checkpoint trained on LIBERO images, without an additional Mobile ALOHA CSD stage. XS-VLA and the SmolVLA-0.25B base use matched policy-training settings and are each trained for 80,000 steps on a single NVIDIA RTX 3090 workstation. Both models are evaluated with the same task instructions, hardware, cameras, and reset procedure; initial object placements are matched as closely as possible.

\textbf{Protocol.}
Each model performs 20 independent trials per task, with robot resets between trials. Cup placing succeeds if the cup remains stably on the plate without falling or being dropped. Sequential coordination requires completing both stages. Block stacking reports strict completed-stack success and a partial-credit score: bringing two blocks together earns one point, and a stable completed stack earns two points total, for at most 40 points over 20 trials. Average task success is the unweighted mean of the three strict success rates; the partial-credit score is reported separately.

\textbf{Results.}
XS-VLA improves all three tasks, raising average strict success from 21.7\% to 65.0\% (Table~\ref{tab:real_world_results_enriched}). Stacking improves from 0\% to 30\% strict success and from 3/40 to 20/40 partial credit, while remaining the most challenging task. The gains span single-arm placement, bimanual stacking, and sequential coordination, extending the combined framework's control benefits to physical manipulation.

Additional deployments on OpenARM, PiPER, and XLeRobot are described in the Appendix.

\subsection{Deployment Footprint and Inference Rate}
\label{sec:deployment_profile}
The full XS-VLA Huber checkpoint at 160,000 steps contains 244,826,880 parameters. Removing 840,544 training-only parameters, including the posterior encoder, leaves 243,986,336 deployed parameters (243.99M), including the action expert.

We profile batch-one inference with two $256\times256$ images, BF16, and 10 flow steps, using eight warm-up runs and 50 measurements. Tokenization, environment simulation, and video writing are excluded. Reported peak inference VRAM is 0.51\,GiB on both GPUs. A 50-action chunk takes 259.93\,ms on A100 PCIe 40GB (3.85 queries/s) and 321.08\,ms on RTX 3090 (3.11 queries/s).

LIBERO executes 10 of the 50 predicted actions before replanning. For chunk latency $L$, the model query rate is $1/L$, while $10/L$ gives amortized rates of 38.47 and 31.15 actions/s on A100 and RTX 3090, respectively. These model-only rates exclude execution overhead and do not measure the robot's servo frequency.

\subsection{Auxiliary Spatial Diagnostics}
\label{sec:diagnostics}
We examine the step-14,000 CSD checkpoint before downstream policy training. Teacher-label agreement measures learning of the auxiliary task, while instruction-pair tests examine how region predictions respond to a change in the target instruction.

\textbf{Teacher-label agreement.}
A human audit finds 36/50 teacher labels correct (72.0\%), with ambiguous target or boundary cases counted as incorrect unless clearly matched. From 5,313 frames in the 20 held-out episodes, we sample 500 unique frames uniformly (seed 42), covering eight region classes. A balanced set (seed 43) contains 100 frames per supported class: bottom center, bottom right, middle left, middle right, and center.

The default VLM and CSD checkpoint use identical inputs and nine-label prompts, greedy decoding, at most 16 new tokens, and no test-time augmentation. A shared parser extracts the initial recognized label. Micro accuracy counts frames; macro accuracy averages observed classes. Both use the teacher-generated region labels as reference targets.

\begin{table}[t]
\centering
\caption{Auxiliary spatial diagnostics (\%) for the 14k checkpoint. Top: teacher-label agreement; macro covers eight natural or five balanced classes. Bottom: instruction-pair tests. Region counts individual inputs; Switch and Pair count pairs.}
\label{tab:spatial_diagnostics}
\small
\begin{tabular}{lrrrr}
\toprule
& \multicolumn{2}{c}{Natural} & \multicolumn{2}{c}{Balanced}\\
\cmidrule(lr){2-3}\cmidrule(lr){4-5}
Model & Micro & Macro & Micro & Macro\\
\midrule
Default SmolVLM2 & 0.0 & 0.0 & 0.0 & 0.0\\
Majority baseline & 39.0 & 12.5 & 20.0 & 20.0\\
CSD student & \textbf{53.4} & \textbf{60.7} & \textbf{58.6} & \textbf{58.6}\\
\bottomrule
\end{tabular}

\medskip
\setlength{\tabcolsep}{3pt}
\begin{tabular}{lrrr}
\toprule
Instruction-pair test & Region & Switch & Pair\\
\midrule
Cross-category & 26.0 & 82.0 & 10.0\\
Seen-noun, fixed template & 50.0 & 12.2 & 2.4\\
\bottomrule
\end{tabular}
\end{table}

CSD achieves 267/500 correct natural-set and 293/500 balanced-set predictions, exceeding majority baselines (Table~\ref{tab:spatial_diagnostics}). Under this prompt, the default model always predicts \texttt{top left}, a class absent from the sampled sets, and receives a zero score. Natural-set \texttt{center} frequency is 195/500; three rare classes have only two examples each. CSD retains a lower-center bias, assigning 85/195 natural-set center examples to bottom~center.

\textbf{Instruction-pair tests.}
The cross-category diagnostic contains 50 unique images, each paired with two target instructions, yielding 50 pairs and 100 inputs. The stricter seen-noun diagnostic retains 41 unique images from 91 candidates, yielding 41 pairs and 82 inputs with a fixed instruction template. Region accuracy evaluates individual inputs; switch rate counts pairs with changed region predictions; pair accuracy requires both predictions to be correct.

Cross-category results are 26/100 correct inputs, 41/50 switched predictions, and 5/50 fully correct pairs. Seen-noun results are 41/82 correct inputs, 5/41 switches, and 1/41 fully correct pairs (Table~\ref{tab:spatial_diagnostics}). The cross-category test shows frequent responses to an instruction change but limited paired accuracy. In the fixed-template test, most region predictions remain unchanged. Each test characterizes its own sample composition; the difference between them cannot be attributed solely to noun familiarity.

CSD improves downstream policy success despite limited accuracy on these auxiliary localization tests. This pattern supports its role as representation initialization for control, with robot task performance providing the primary measure of its utility.

\section{Discussion and Limitations}
\label{sec:limitations}
The results connect CSD initialization and regularized demonstration-conditioned flow learning to improved compact control. Further study can clarify which properties of the learned representations and latent training account for these~gains.

\textbf{Latent/KL attribution and prior-mean inference.}
The current branch ablation changes posterior conditioning and KL together. A matched run retaining the posterior with $\lambda_{\mathrm{KL}}=0$ would help distinguish their contributions through comparison with full LFM and the no-posterior control. Posterior statistics and zero-versus-sampled-latent evaluations could further clarify how regularization relates to prior-mean inference; latent informativeness and motion-style disentanglement have not been established.

\textbf{Representation initialization and control.}
The CSD diagnostics show limited paired-target localization in the presence of teacher noise, coarse labels, and position bias. We therefore interpret CSD as initialization for downstream control, with reliable instruction-conditioned grounding still an open question. Identifying which representation properties drive the gains, and how they evolve during joint policy training, calls for analysis beyond the generic-text pretraining~control.

\textbf{Augmentation consistency.}
Horizontal reflection with label swapping preserves the annotated target's image-region correspondence. Instructions whose left/right relations determine target selection may also require corresponding text transformations. We have not evaluated that semantic consistency or a no-flip control; both would help assess the role of augmentation in CSD.

\textbf{Experimental scope.}
The full model has two training-seed evaluations, while the controls have one each. Broader seed coverage would quantify variability in the observed differences. Physical evaluation covers three Mobile ALOHA tasks with 20 trials per model and task; the intervals describe trial uncertainty. Component ablations on hardware would help assess each stage's contribution. Additional deployments are qualitative, and the GPU profile characterizes XS-VLA alone. Baseline profiling and end-to-end timing would extend the deployment comparison.

\section{Conclusion}
XS-VLA learns compact robot control in two stages: coarse spatial supervision initializes the backbone, and KL-regularized demonstration latents condition flow matching. The 243.99M-parameter deployed policy achieves 90.25\% LIBERO success in two training seeds and 65.0\% average strict success on Mobile ALOHA, while operating without the teacher or posterior encoder. The control gains across simulation and physical manipulation demonstrate the value of richer training supervision within a compact deployment~footprint.

\appendix[Deployment on Additional Robot Platforms]
\label{app:deployments}
We demonstrate XS-VLA on OpenARM, PiPER, and XLeRobot through platform-specific deployments. The following examples provide qualitative evidence of operation beyond the main Mobile ALOHA evaluation.

\begin{figure*}[t]
\centering
\includegraphics[width=0.95\textwidth]{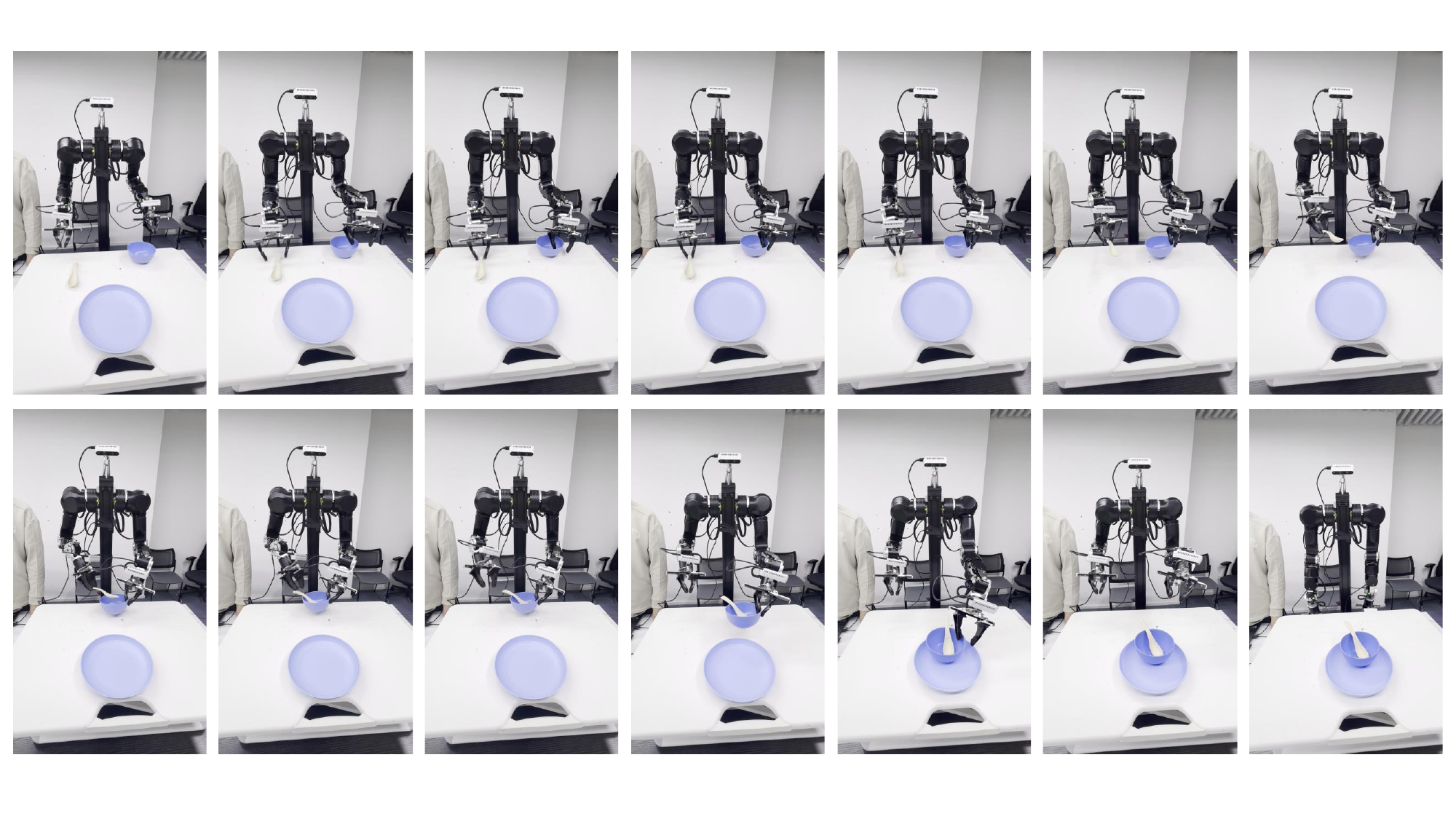}
\caption{XS-VLA operating on OpenARM during table tidying. The frame sequence illustrates operation on a robot platform beyond the main Mobile ALOHA evaluation.}
\label{fig:openarm}
\end{figure*}

\textbf{OpenARM.}
Figure~\ref{fig:openarm} shows XS-VLA performing table tidying on OpenARM.

\begin{figure}[ht]
\centering
\includegraphics[width=\columnwidth]{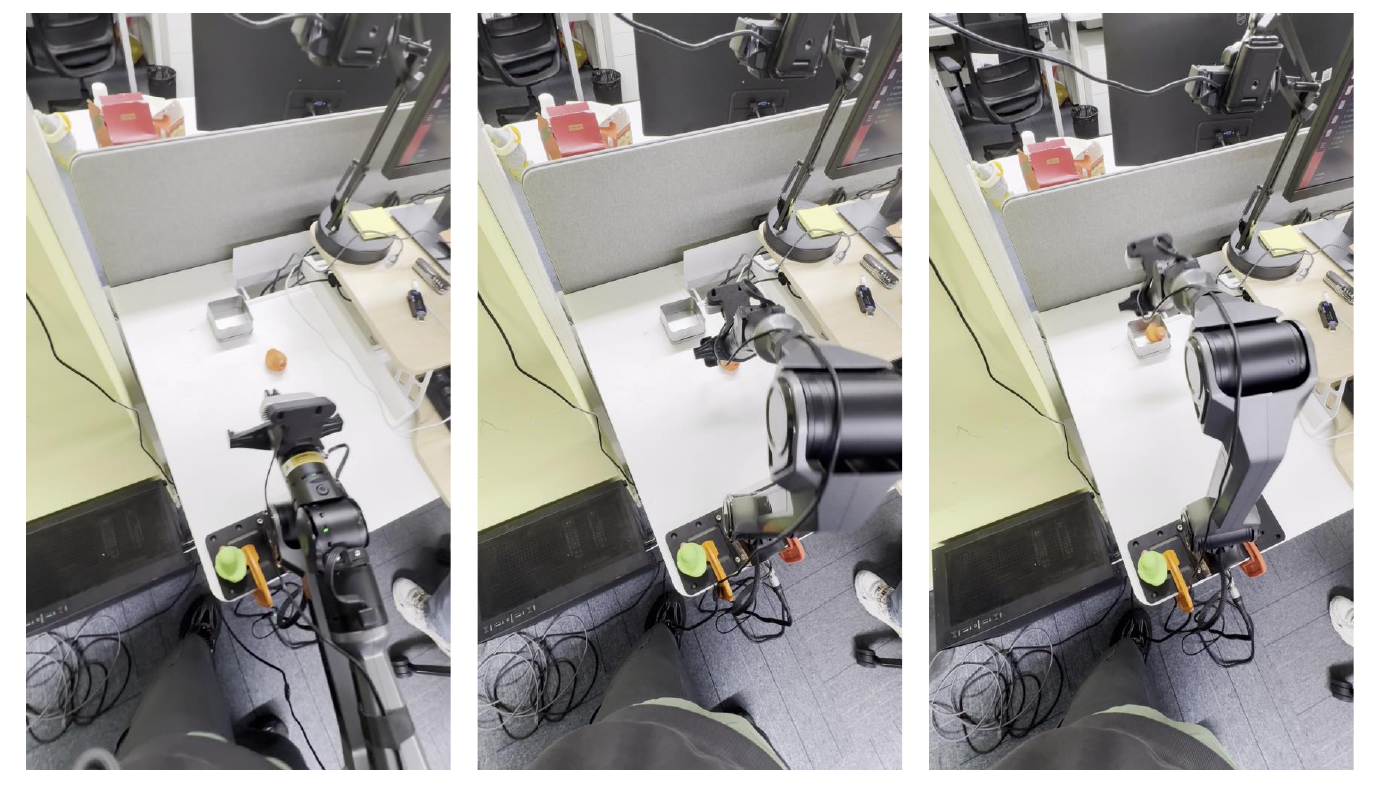}
\caption{XS-VLA operating on PiPER: successive frames show the robot picking up an orange duck. Together with OpenARM, this sequence provides qualitative evidence of deployment across robot platforms.}
\label{fig:piper}
\end{figure}

\textbf{PiPER.}
Figure~\ref{fig:piper} shows XS-VLA picking up an orange duck on PiPER, providing another example of operation on a different robot platform.

\textbf{XLeRobot.}
We also deployed XS-VLA on XLeRobot~\cite{wang2025xlerobot} for a bimanual carrot-transfer task: the left arm picks up the carrot and transfers it to the right arm for placement. This deployment uses 100 task demonstrations collected by three~operators.

\begingroup
\setlength{\emergencystretch}{1em}
\bibliographystyle{IEEEtran}
\bibliography{main}
\endgroup
\end{document}